\documentclass[10pt,twocolumn,a4paper]{article}

\usepackage[final]{cvwwarxiv}

\usepackage{graphicx}
\usepackage{amsmath}
\usepackage{booktabs}
\usepackage{tabularx}
\usepackage{makecell}
\usepackage{multirow}
\usepackage{array}

\usepackage{tikz}
\usepackage{stfloats} 
\usepackage[dvipsnames]{xcolor}
\usetikzlibrary{shapes.geometric,arrows.meta}

\usepackage[pagebackref,breaklinks,colorlinks,allcolors=cvwwblue]{hyperref}

\def\paperID{45}

\title{Grading Handwritten Engineering Exams with Multimodal Large Language Models}

\author{Janez Perš, Jon Muhovič, Andrej Košir and Boštjan Murovec\\
University of Ljubljana, Faculty of Electrical Engineering\\
{\tt\small \{janez.pers\},\{jon.muhovic\},\{andrej.kosir\},\{bostjan.murovec\}@fe.uni-lj.si}
}

\begin{document}
\maketitle

\begin{abstract}
Handwritten STEM exams capture open-ended reasoning and diagrams, but manual grading is slow and difficult to scale. We present an end-to-end workflow for grading scanned handwritten engineering quizzes with multimodal large language models (LLMs) that preserves the standard exam process (A4 paper, unconstrained student handwriting). The lecturer provides only a handwritten reference solution (100\%) and a short set of grading rules; the reference is converted into a text-only summary that conditions grading without exposing the reference scan. Reliability is achieved through a multi-stage design with a format/presence check to prevent grading blank answers, an ensemble of independent graders, supervisor aggregation, and rigid templates with deterministic validation to produce auditable, machine-parseable reports. We evaluate the frozen pipeline in a clean-room protocol on a held-out real course quiz in Slovenian, including hand-drawn circuit schematics. With state-of-the-art backends (GPT-5.2 and Gemini-3 Pro), the full pipeline achieves $\approx$8-point mean absolute difference to lecturer grades with low bias and an estimated manual-review trigger rate of $\approx$17\% at $D_{\max}=40$. Ablations show that trivial prompting and removing the reference solution substantially degrade accuracy and introduce systematic over-grading, confirming that structured prompting and reference grounding are essential.
\end{abstract}

\section{Introduction}
\begin{figure}[t]
  \centering
  \includegraphics[width=\linewidth]{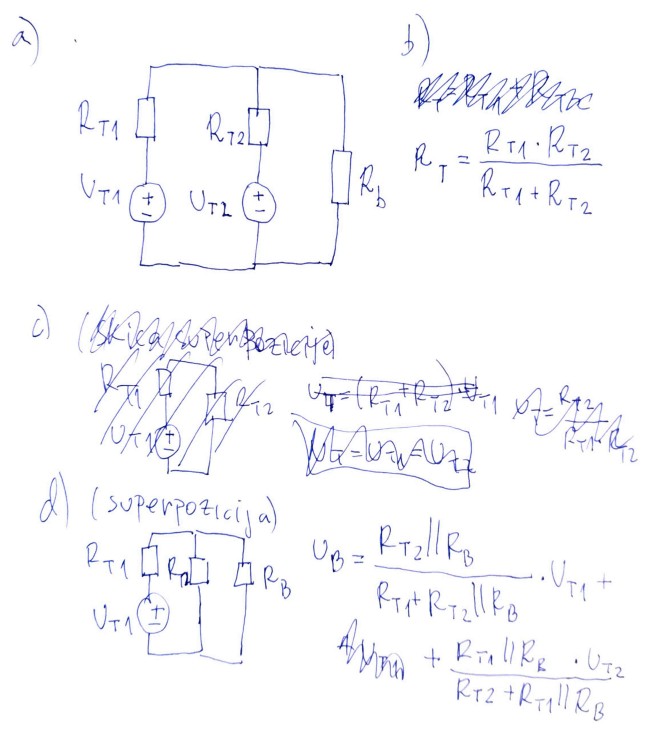}
  \caption{Sample handwritten answer with diagrams (e.g.\ circuits) that our system is designed to handle. Sample provided by authors (not student work). The link to the full exam and its grading result are provided as supplementary material in the last section of the paper.}
  \label{fig:examsample}
\end{figure}

Handwritten, paper-based exams remain common in STEM education because they naturally elicit open-ended reasoning, intermediate work, and sketches (e.g., circuit diagrams) that are difficult to capture with purely digital assessments. Figure~\ref{fig:examsample} illustrates the challenges. Yet manual grading of such exams is time-consuming and hard to scale. Learnosity reports that, in an online survey of 258 U.S.\ teachers, respondents spent an average of 9.9 hours per week grading and marking \cite{Learnosity2025}. At the university level, faculty emotions research similarly suggests that grading can be experienced as comparatively unpleasant: in a large U.S.\ faculty sample, grading was associated with less positive and more negative emotions than research and teaching \cite{Schwab2023}. Beyond the immediate time cost, practitioner-facing syntheses argue that extensive grading can consume time and energy that would otherwise support course planning and instructional improvement \cite{Terada2024}.

The problem is amplified in STEM courses, where written exams and technical assignments require evaluating multi-step solutions and graphical artifacts under time pressure. A STEM-focused review notes that, as enrollments and workloads grow, instructors can struggle to provide timely feedback on labor-intensive assessments and may assign fewer practice opportunities despite pedagogical benefits \cite{Tan2025}. More broadly, automated grading systems are often motivated by reducing educator workload and shortening feedback turnaround \cite{Tan2025}. Workflow systems such as Gradescope show that digitizing and organizing scanned handwritten submissions can reduce grading overhead and support more consistent rubric application \cite{Singh2017gradescope}, but these platforms largely preserve a fundamental bottleneck: humans still read and score each response.

Recent progress in large language models (LLMs) opens a path toward more automated grading. On text-only short-answer grading, experiments on K--12 short-answer data report that GPT-4 can achieve agreement close to human rater agreement under relatively simple prompting \cite{Henkel2024LLMgrade}. Controlled ASAG evaluations further show that grading performance is sensitive to prompt context---including whether a reference answer is provided or withheld---and can change substantially across datasets and setups \cite{Kortemeyer2024GPT4ASAG}. For paper-based STEM exams, multimodal LLMs can be evaluated directly on images of handwritten solutions; recent results report improved alignment when prompts include official solutions and a grading rubric, while also noting that accuracy can remain insufficient for unsupervised, real-world deployment \cite{Caraeni2024GPT4vision}. Practical limitations also remain in hybrid workflows: when handwriting is transcribed to text/\LaTeX{} before grading, transcription errors can affect downstream scoring, and recent work treats LLM-produced grades as subject to subsequent human verification and explores confidence estimation via repeated sampling \cite{Liu2024AIassistedHandwrittenMathGrading}. Separately, applied deployments on longer, multi-part responses highlight handwriting-to-text conversion as a recurring bottleneck and report challenges in applying fine-grained rubrics to long solutions and diagram/graph-heavy work \cite{Kortemeyer2024thermo}. Together, these findings suggest that reliable exam grading requires not only strong multimodal models, but also systems-level design: structured prompting, verification/aggregation, and deterministic post-processing to enforce consistent outputs.

In this work, we target the setting of end-to-end grading for scanned handwritten STEM exams that combine free-form text with hand-drawn diagrams. We present a multi-stage, multi-prompt grading workflow with deterministic post-processing to ensure reliable, auditable outputs, and evaluate it on undergraduate open-question engineering exams requiring handwritten textual answers and electrical schematics.

\section{Related Work}
Research into automated exam grading has a long history that predates modern LLMs. Before LLMs, many practical systems either (i) constrained answers into computer-readable formats (enabling direct scoring) or (ii) relied on handwriting recognition/OCR to extract text from scanned work before applying text-based scoring methods. A recurring challenge in this area is data availability: authentic exam scripts are frequently private and difficult to share, and publicly distributable datasets often only partially reflect real exam conditions.

\subsection{Exam grading before the advent of LLMs}
Prior to multimodal LLMs, automatic assessment largely decomposed into (i) text-based automated short-answer grading (ASAG) and (ii) document pipelines that first transcribed handwriting. Surveys of ASAG describe early systems built around engineered lexical/syntactic features and semantic similarity to reference answers, often using supervised models trained on scored responses and evaluated with standard agreement/correlation metrics \cite{Burrows2015eras}. Representative feature-integration approaches combine multiple linguistic feature families in discriminative scoring models \cite{Sakaguchi2015short}, while vector-based approaches use distributed representations and similarity scoring for grading \cite{Magooda2016vector}. More recent surveys emphasize the shift toward deep learning and pretrained language models (including transformer-based approaches) for text ASAG \cite{Haller2022survey}.

For handwritten work, a common strategy was handwriting recognition/OCR followed by text-based scoring. Early examples integrated handwriting recognition with automated essay scoring and underscored the dependence of end-to-end scoring quality on transcription quality \cite{Srihari2007handwritten}. For short handwritten answers, Rowtula et al.\ propose a word-spotting-based approach that avoids full transcription and instead relies on retrieval-style signals for downstream evaluation \cite{Rowtula2019handwritten}. In parallel, system-level tools such as Gradescope scaled the \emph{workflow} of grading scanned submissions through dynamic rubrics and answer grouping, while still relying on humans to assign points \cite{Singh2017gradescope}. For structured STEM problems, clustering-based methods have been proposed to group similar solution structures so that an instructor can label clusters and propagate (partial) credit \cite{Lan2015MLP}. Overall, pre-LLM grading systems either assumed clean text or relied on transcription as an error-prone front-end, and most did not robustly support unconstrained handwriting, diagrams, and multi-page exam context end-to-end.

\subsection{Exam grading in the LLM era}
Since 2023, large language models have been evaluated as general-purpose graders, first on text-only responses and increasingly on images of handwritten work. On K--12 short-answer data, Henkel et al.\ report that GPT-4 with relatively simple prompting can achieve agreement close to human rater agreement \cite{Henkel2024LLMgrade}. Complementary ASAG studies show that results depend on prompt context and experimental setup, including whether reference answers are provided or withheld \cite{Kortemeyer2024GPT4ASAG}. In advanced mathematics settings, Gandolfi reports that GPT-4 can produce useful solutions and grading rationales, while also documenting reliability issues such as occasional loss of coherence and hallucinations that motivate explicit verification and guardrails \cite{Gandolfi2024GPT4education}.

A major shift is direct multimodal grading of scanned handwritten solutions. For university-level math exams, Caraeni et al.\ evaluate GPT-4o grading directly from handwriting images and report improved alignment when prompts include official solutions and a grading rubric, while noting that overall accuracy remains a limiting factor for real-world use \cite{Caraeni2024GPT4vision}. Hybrid pipelines remain relevant: Liu et al.\ study AI-assisted grading of handwritten university mathematics exams using an OCR/\LaTeX{} transcription stage and emphasize both transcription sensitivity and the role of human verification, while also exploring confidence estimation by sampling multiple grading runs \cite{Liu2024AIassistedHandwrittenMathGrading}. At larger scale (e.g., hundreds of scripts in physics), workflow studies continue to highlight handwriting-to-text conversion as a practical bottleneck, challenges in applying fine-grained rubrics to long solutions, and persistent difficulty with diagrams/graphs \cite{Kortemeyer2024thermo}. For non-textual outputs such as hand-drawn graphs, recent comparative work evaluates meta-learning approaches alongside vision-language models on graph grading tasks \cite{Parsaeifard2025graphs}. Alternative multimodal scoring paradigms use CLIP-style embeddings (optionally combined with OCR) to incorporate visual information into scoring \cite{Baral2023AutoScoringImagesMath}, and new benchmarks (e.g., DrawEduMath) systematically probe VLM interpretation of students' hand-drawn math images and document remaining weaknesses \cite{Baral2025DrawEduMath}.

\begin{figure*}[!ht]
  \centering
  \begin{tikzpicture}[scale=0.90, transform shape, font=\scriptsize]

  \tikzset{
    processColor/.style={draw=MidnightBlue, text=MidnightBlue, fill=MidnightBlue!5},
    dataColor/.style={draw=ForestGreen, text=ForestGreen, fill=ForestGreen!5},
    promptColor/.style={draw=BrickRed, text=BrickRed, fill=BrickRed!5},
    process/.style={
      rectangle, rounded corners, processColor,
      line width=1.2pt, minimum width=3.2cm, minimum height=1.2cm, align=center
    },
    data/.style={
      rectangle, dataColor,
      line width=1.2pt, minimum width=3.2cm, minimum height=0.9cm, align=center
    },
    ellip/.style={
      ellipse, promptColor,
      line width=1.2pt, minimum width=3.4cm, minimum height=1.0cm, align=center
    },
    arrowBase/.style={-Latex, line width=1.2pt},
    arrowProcess/.style={arrowBase, draw=MidnightBlue},
    arrowData/.style={arrowBase, draw=ForestGreen},
    arrowPrompt/.style={arrowBase, draw=BrickRed}
  }

  \node[data]    (student)    at (0,1.5)   {Pseudo-anonymized\\scanned student exam};
  \node[process] (format)     at (4,1.5)   {Format /\ presence\\checker};
  \node[process] (grading)    at (8,1.5)   {Grading engine\\(ensemble graders)};
  \node[process] (supervisor) at (12,1.5)  {Supervisor\\aggregation};
  \node[process] (postproc)   at (16,1.5)  {Postprocessor};
  \node[data]    (outputs)    at (16,-0.8) {Final PDFs,\\reports, exports};

  \node[ellip] (fmtPrompt)    at (4,3.2)   {Format / presence\\prompt pair};
  \node[ellip] (gradePrompt)  at (8,3.2)   {Grading\\prompt pair};
  \node[ellip] (supPrompt)    at (12,3.2)  {Supervisor\\prompt pair};
  \node[ellip] (postPrompt)   at (16,3.2)  {Postprocessor\\prompt pair};

  \node[ellip] (templates) at (11.2,-1.0)
    {Markdown templates\\\& grading/report layout};

  \node[ellip] (examids) at (6.0,0.0) {Pseudoanonymous\\student ID list};
  \draw[arrowPrompt] (examids) -- (grading.south);

  \draw[arrowData]    (student)    -- (format);
  \draw[arrowProcess] (format)     -- (grading);
  \draw[arrowProcess] (grading)    -- (supervisor);
  \draw[arrowProcess] (supervisor) -- (postproc);
  \draw[arrowProcess] (postproc)   -- (outputs);

  \draw[arrowPrompt] (fmtPrompt)   -- (format);
  \draw[arrowPrompt] (gradePrompt) -- (grading);
  \draw[arrowPrompt] (supPrompt)   -- (supervisor);
  \draw[arrowPrompt] (postPrompt)  -- (postproc);

  \draw[arrowPrompt] (templates) -- (grading.south);

  \node[data]    (refscan) at (0,-3.0)  {Scanned\\reference exam};
  \node[process] (refproc) at (4,-3.0)  {Reference solution\\extraction};
  \node[data]    (reftext) at (8,-3.0)  {Reference solution\\summary (text)};

  \node[ellip] (refPrompt) at (4,-1.0)
    {Reference solution\\prompt pair};

  \draw[arrowData]    (refscan) -- (refproc);
  \draw[arrowProcess] (refproc) -- (reftext);
  \draw[arrowPrompt]  (refPrompt) -- (refproc);

  \draw[arrowData] (reftext) -- ++(0,2.0) -| (grading.south);

  \end{tikzpicture}
  \caption{System overview. Green boxes denote data artifacts, red ellipses denote prompt pairs and templates, and blue boxes denote processing stages. All blue processing stages invoke a multimodal LLM backend (e.g., GPT-4o, GPT-5.x, Gemini, or Mistral), while reference solution scans are converted into a text-only summary that is injected into grading prompts.}
  \label{fig:overall}
  \vspace{-1.0em}
\end{figure*}
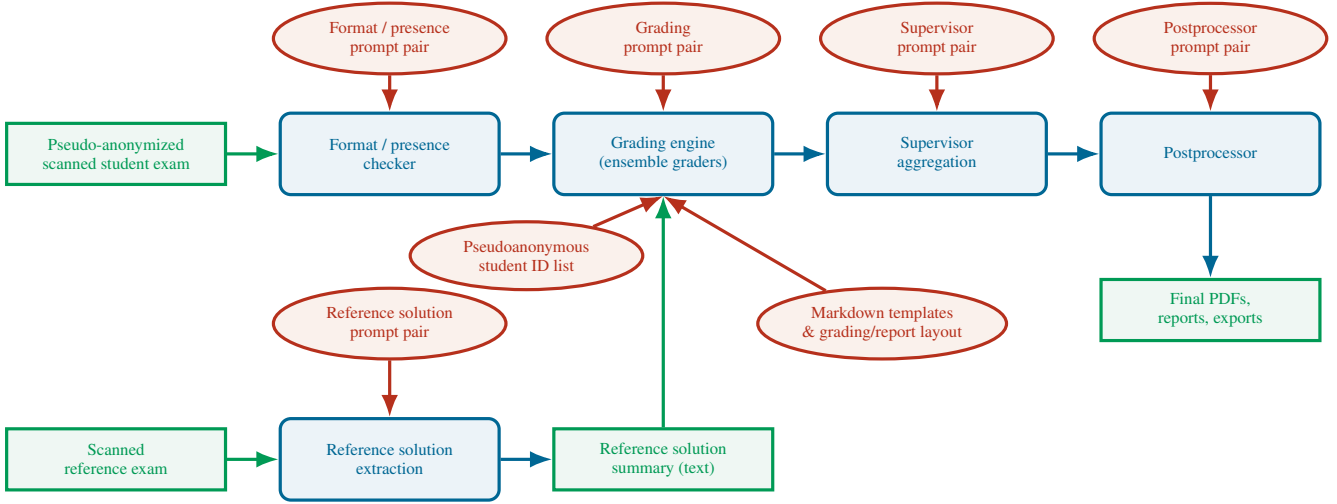

\subsection{Datasets}
Public datasets that enable \emph{end-to-end} evaluation of handwritten exam grading (images in, numeric scores out) are rare, largely because authentic exam scripts are privacy-sensitive and typically collected under institutional consent/IRB, limiting public release. Consequently, recent multimodal grading studies evaluate on private course-exam collections with scanned pages, rubrics, and human scores \cite{Caraeni2024GPT4vision,Kortemeyer2024thermo}, while graph-focused handwritten corpora are likewise institutional \cite{Parsaeifard2025graphs}. The few open resources with handwriting either target short answers \cite{GoldZesch2020HandwrittenASAP} or serve as VLM understanding benchmarks rather than points-based grading datasets \cite{Baral2025DrawEduMath}; large educational corpora with image responses are often not redistributable \cite{Baral2021ImprovingAutoScoringMathOpenResponses}. Therefore, to evaluate an end-to-end pipeline under realistic STEM exam conditions (multi-part solutions and diagrams), we collected and graded our own exam data.


\section{Methods}

The pipeline was designed to minimize interference with the exam process. Beyond using the standard A4 paper, no constraints are imposed on students or on how the lecturer administers the exam. The only additional requirement is that the lecturer provides a handwritten reference solution representing a perfect (100\%) answer. These requirements directly shape the architecture shown in Fig.~\ref{fig:overall}.

\subsection{System architecture}
\label{sec:system_arch}
The pipeline in Fig.~\ref{fig:overall} couples deterministic document handling with a small number of LLM-invoking stages. While the system is implemented as a substantial amount of orchestration code, the primary methodological contribution lies in the \emph{structure of the pipeline} and, crucially, in the \emph{prompt pairs and rigid templates} that make the overall behavior stable and machine-parseable. Due to space constraints, we do not reproduce the full prompts and templates.

\paragraph{Reference conditioning without exposing the reference scan.}
The lecturer provides a handwritten reference solution representing a perfect (100\%) answer. A dedicated reference-extraction stage converts the scanned reference into a text-only summary that is injected into grading prompts; the reference image itself is not used during grading. This stage is run with a highly capable multimodal backend (GPT-5.2-pro in our experiments) to robustly interpret unconstrained handwriting and sketches.

\paragraph{Answer presence guardrail.}
Before any scoring, a format/presence checker predicts which tasks contain an actual student answer. This safeguard was introduced after observing rare cases where a model would hallucinate content when an entire answer region was left blank. Although such events are not reflected in aggregate metrics, they are operationally unacceptable, and the presence list prevents the grader from assigning points to missing answers.

\paragraph{Ensemble grading and supervisor aggregation.}
For each task, grading is performed by an ensemble of $K=3$ independent, stateless model calls that produce structured drafts according to a fixed template. A supervisor model then merges the drafts into a single exam-level output, enforces template compliance, applies the presence decisions (unanswered tasks receive 0\%), and flags inconsistencies for optional human resolution. This ensemble-plus-supervisor design reduces variance and improves robustness to occasional model glitches.

\paragraph{Postprocessing and exports.}
Finally, a postprocessor produces presentation-ready artifacts (e.g., report assembly and optional translation) while preserving all numeric fields and the template structure, enabling deterministic parsing, auditing, and export.

\paragraph{Privacy.}
Students are instructed \emph{not to write their name and surname anywhere on the exam.} Instead, they are required to write their registration number, so the data is pseudo-anonymized before invoking LLM. The LLM gets sanitized \emph{student roster} with all registration numbers for students participating in the exam and only needs to find which one of the finite set of registration number is written at the top of the first page of the exam. All de-anonymization is done locally after grading. Mandatory markdown templates ensure that the model outputs student's pseudo-identity in a specific place in the output, where it can be parsed. If parsing fails, human is required to read the number from the scan.

\subsection{Prompts and templates}
\label{sec:prompts_templates}
Each LLM-invoking stage is implemented as a \emph{prompt pair} (system + user) backed by rigid Markdown templates. The system prompt fixes role, constraints, and prohibited behavior, while the user prompt injects instance-specific context (task labels, student scan, reference summary, and optional instructor rules). The templates strictly define section hierarchy and numeric fields, converting otherwise probabilistic outputs into artifacts that can be validated and parsed by deterministic post-processing; in contrast, directly asking an LLM to ``grade the exam'' is unreliable due to format drift and inconsistent application of rubrics.

\paragraph{Instructor-facing configuration.}
The only course-specific inputs are (i) the scanned handwritten reference solution and (ii) a short list of grading rules. All other pipeline stages, prompts, and templates are intended to remain unchanged across courses. Rules are numbered (\texttt{[R1]}, \texttt{[R2]}, \dots), appended verbatim to grading-related prompts, and graders are instructed to cite applicable rule IDs in their explanations, improving auditability and facilitating human review. In our experiments, most rules transferred across STEM quizzes, with only minor course-specific adjustments (e.g., evaluating circuit sketches by topology rather than drawing orientation).

\paragraph{Template constraints.}
Two templates are used: a per-grader template and a supervisor template. Both prohibit adding or removing sections and enforce a fixed scoring line with a deterministic numeric pattern (achievement, weight, contribution). Per task, the template separates the question text, a plain-text summary of the student answer, a correctness explanation with required short meta-tags (including rule citations), and a single scoring line. The final exam total must equal the sum of per-task contributions (no normalization), which enables automatic consistency checks.

\paragraph{Language.}
Unless stated otherwise, the prompts, rules, and templates used in our experiments are written in Slovenian. The pipeline itself is language-agnostic: adapting to other languages requires only translating these text artifacts, without changing the processing stages.

\section{Experiments}
\label{sec:experiments}

Automated grading quality has important qualitative aspects that are not fully captured by aggregate metrics alone (e.g., the coherence and pedagogical usefulness of the generated feedback). Therefore, in addition to the quantitative evaluation reported below, we provide \emph{supplementary material}, comprised of the scanned reference solution of the "Class B" exam, and a corresponding mock solution, which was actually graded by the pipeline using GPT-5.2 with thinking set to "high". We provide these materials to illustrate the structure and depth of the produced feedback. The link to the supplementary material is provided at the end of the manuscript.

\subsection{Experimental protocol}

Our grading pipeline is not trainable and therefore operates in a fully zero-shot setting. To limit potential bias from iterative prompt and rubric engineering, we use a clean-room protocol with two parallel courses: \textbf{Class A} (development) and \textbf{Class B} (held-out evaluation).

During the semester, the system was deployed weekly on Class~A quizzes and its outputs were reviewed by the lecturer and students, providing formative feedback on system behavior. After week~9, the pipeline and all prompts were frozen. The frozen system was then applied to one scanned quiz from Class~B \emph{without any modifications to the pipeline, prompts, or templates}. The only change relative to Class~A was an instructor-facing grading-rule adjustment reflecting standard practice for circuit sketches: circuit answers were evaluated by electrical topology rather than drawing orientation. All quantitative and qualitative results reported in this paper are based exclusively on this held-out Class~B evaluation and are compared to grades assigned by the Class~B lecturer.

To maintain separation, Class~B quizzes were graded only by the lecturer during the semester and scanned for archival purposes. Apart from limited pilot runs in week~2 (not used for analysis), the Class~B materials were not accessed or inspected by the system developers prior to the held-out evaluation.

\subsection{Dataset}

Class~A is an undergraduate course on communication technologies (approximately 30 enrolled students), while Class~B is an industrial electronics course (approximately 15 enrolled students). Both courses used short weekly written quizzes administered for 20 minutes at the beginning of each lecture, covering material from the previous week. The quizzes consist of open-ended questions requiring handwritten text and, where appropriate, hand-drawn diagrams or schematics.

Weekly quizzes were not strictly mandatory, but passing a subset was required to qualify for the final exam; strong weekly performance could optionally substitute the final exam grade. This provided meaningful incentive while keeping overall pressure moderate. For privacy reasons, we do not release student submissions, scans, or grades; we report only aggregate performance metrics and the exact text of the held-out evaluation questions.

\paragraph{Language.}
The quizzes and student answers were in a non-English language (Slovenian). Accordingly, all prompt pairs, grading rules, and report templates used in the main evaluation were written in Slovenian. The pipeline itself is language-agnostic: language-specific content is confined to prompts, rules, and templates, which can be translated to other languages (including with the assistance of modern LLM-based tools) without changing the pipeline stages.

\paragraph{Held-out exam content.}
Table~\ref{tab:exam_questions} lists the three questions (with weights) from the held-out Class~B quiz used for evaluation.

\begin{table}[t]
\centering
\small
\begin{tabular}{p{0.10\linewidth} p{0.80\linewidth}}
\hline
\textbf{Weight} & \textbf{Question text} \\
\hline
25\% &
When using voltage dividers, we encounter a trade-off: for certain reasons we want to construct the divider using resistors with as small resistance values as possible, while on the other hand we want the resistances to be as large as possible. Explain this contradiction and the reasons behind it. \\[0.6em]

25\% &
What condition must be satisfied when connecting a load to a voltage divider consisting of resistors $R_1$ and $R_2$ in order for the load to be current-driven? \\[0.6em]

50\% &
Two batteries with Thevenin voltages $U_{t1}$ and $U_{t2}$ and Thevenin internal resistances $R_{t1}$ and $R_{t2}$ are connected in parallel and then connected to a load $R_b$.
\begin{itemize}\setlength{\itemsep}{0pt}
  \item Sketch the corresponding circuit.
  \item Write the expression for the Thevenin resistance of the combined source.
  \item Write the expression for the Thevenin voltage of the combined source.
  \item Write the expression for the voltage across the load.
\end{itemize} \\
\hline
\end{tabular}
\caption{Exam questions used in the held-out evaluation. Only question text and weights are disclosed; no student data are shared. Questions are translated from Slovenian; translation is provided for readability.}

\label{tab:exam_questions}
\end{table}

\subsection{Evaluation Metrics}
\label{sec:eval_metrics}
We evaluate the grading system using exam-level metrics that quantify agreement with human grading. As ground truth, we use the exam grade assigned by the lecturer of Class~B, which is the only available reference.

Unless stated otherwise, all metrics are computed at the level of the full exam (three tasks), thereby evaluating the system in a true end-to-end setting. Importantly, we distinguish two sources of variability: (i) an internal ensemble used by the grading pipeline as part of its fixed design, and (ii) independent repetitions of the full evaluation used only to estimate experimental variability.

\paragraph{Pipeline parameter: ensemble size.}
Within a single pipeline execution, each task is graded by an ensemble of $K=3$ independent, stateless model calls, and the resulting drafts are merged by a supervisor model into a single exam-level grade. We denote this final, supervisor-aggregated output for student $i$ as $g_i^{\text{LLM}}$. The ensemble size $K$ is a fixed pipeline parameter (not varied in the experiments), chosen early as a compromise between inference cost and robustness to occasional model failures.

Let $N$ denote the number of students, $g_i^{\text{LLM}}$ the exam grade assigned by the system for student $i$, and $g_i^{\text{H}}$ the corresponding grade assigned by the human lecturer. We define the signed grading difference as $\Delta_i = g_i^{\text{LLM}} - g_i^{\text{H}}$.

\paragraph{Mean Absolute Difference (MAD).}
Overall grading accuracy is measured using the mean absolute difference between automated and human-assigned grades:
\begin{equation}
\mathrm{MAD} = \frac{1}{N} \sum_{i=1}^{N} \left| \Delta_i \right|.
\end{equation}

\paragraph{Standard Deviation of Absolute Differences.}
To capture the variability of grading errors across students, we compute the standard deviation of absolute differences:
\begin{equation}
\sigma_{\lvert \Delta \rvert}
= \sqrt{
\frac{1}{N}
\sum_{i=1}^{N}
\left(
\left| \Delta_i \right|
- \mathrm{MAD}
\right)^2
}.
\end{equation}

\paragraph{Grading Bias.}
Systematic over- or under-grading is quantified by the mean signed difference:
\begin{equation}
\mathrm{Bias} = \frac{1}{N} \sum_{i=1}^{N} \Delta_i.
\end{equation}

\paragraph{Manual Review Trigger Rate.}
In addition to agreement with human grades, we quantify the expected amount of manual consolidation required when multiple automated graders disagree. This metric does not rely on human reference grades and is computed solely from the ensemble outputs.

Let $s_{i,k}$ denote the exam-level score assigned to student $i$ by grader $k$ within the ensemble, with $k=1,\dots,K$. For a given disagreement threshold $D_{\max}$, a manual review is triggered for student $i$ if the maximum pairwise disagreement between graders exceeds the threshold, i.e.,
\begin{equation}
\max_{k} s_{i,k} - \min_{k} s_{i,k} \ge D_{\max}.
\end{equation}
We define the corresponding trigger indicator as
\begin{equation}
T_i(D_{\max}) =
\begin{cases}
1, & \text{if a review is triggered for student } i, \\
0, & \text{otherwise}.
\end{cases}
\end{equation}

The manual review trigger rate at threshold $D_{\max}$ is then given by
\begin{equation}
\mathrm{TR}(D_{\max}) = \frac{1}{N} \sum_{i=1}^{N} T_i(D_{\max}).
\end{equation}

In this work, $\mathrm{TR}(D_{\max})$ is estimated at the \emph{exam level}, using final exam scores produced by each grader. In practical deployments, the same criterion could be applied at a finer granularity, such as the level of individual questions or answers, to further localize and reduce the required amount of human intervention.

\paragraph{Experimental parameter: evaluation repetitions.}
Separately from the pipeline ensemble, we repeat the \emph{entire} evaluation $R=3$ times to measure run-to-run variability due to stochastic model outputs. Each repetition corresponds to a full rerun of the grading pipeline (including all model calls and supervisor aggregation) with the system configuration unchanged. For each metric, we report the three per-run values, together with their mean and standard deviation across the $R$ repetitions. The choice $R=3$ reflects a practical compromise between robustness and the computational cost of repeated multimodal inference.

\section{Results}
\begin{figure}[t]
  \centering
  \includegraphics[width=\linewidth]{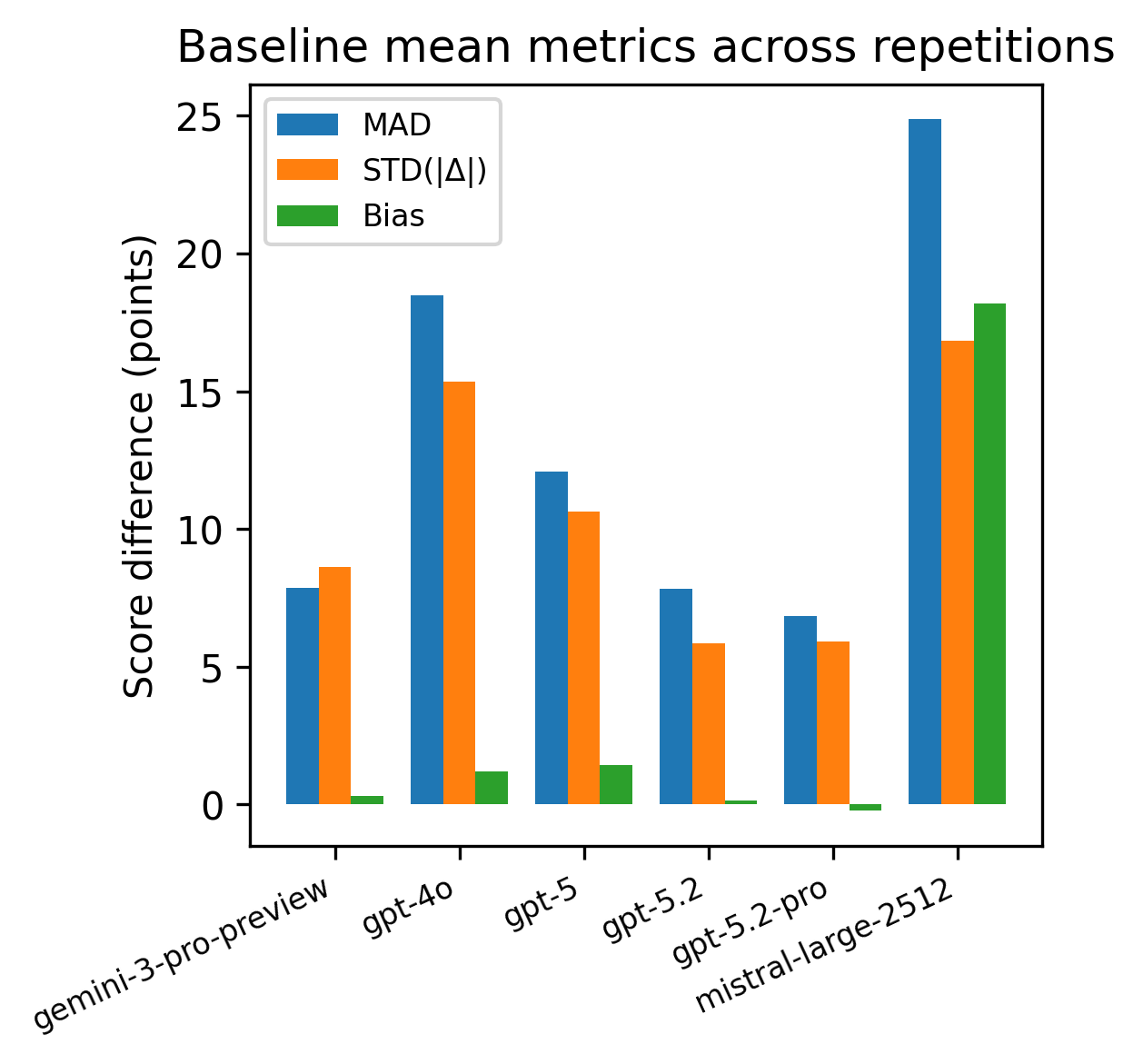}
  \caption{Baseline exam-level performance across backends (mean over $R=3$ repetitions where available; GPT-5.2-pro: single run).}
  \label{fig:result01}
\end{figure}

\begin{figure}[t]
  \centering
  \includegraphics[width=\linewidth]{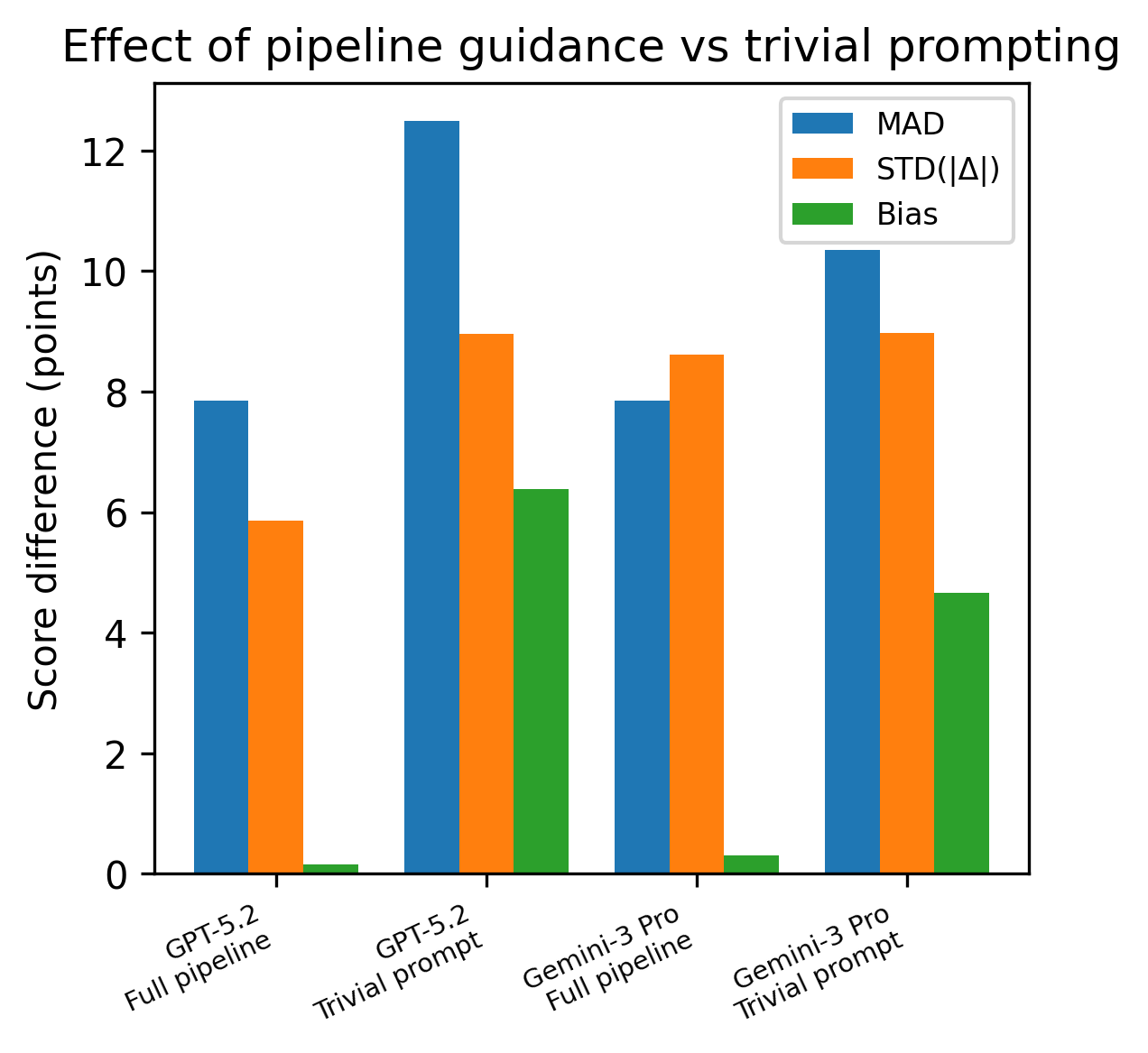}
  \caption{Full pipeline vs.\ trivial prompting for two strong backends (trivial: no rules, no reference, no supervisor).}
  \label{fig:result02}
\end{figure}

\begin{figure}[t]
  \centering
  \includegraphics[width=\linewidth]{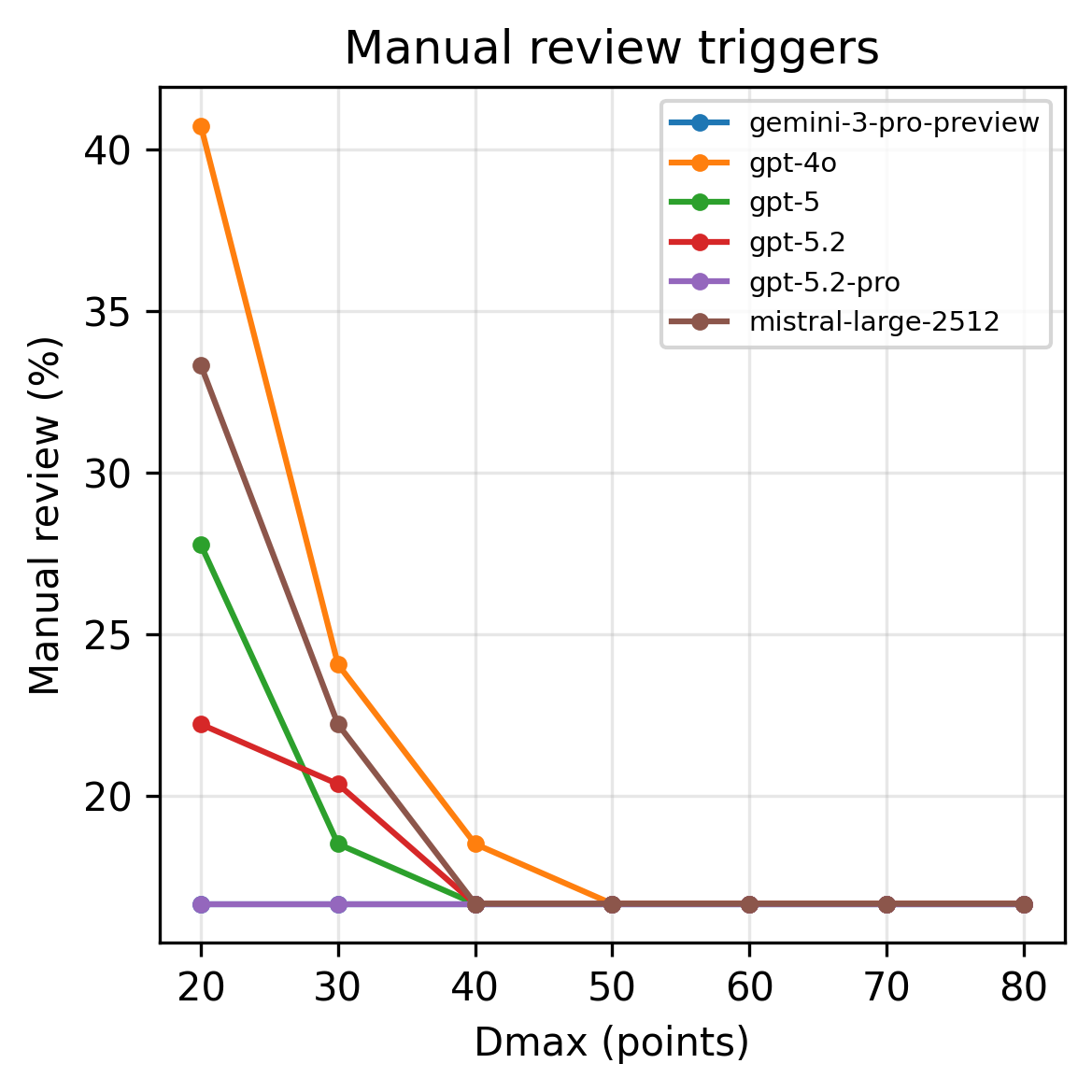}
  \caption{Estimated manual review trigger rate vs.\ disagreement threshold $D_{\max}$ (mean over repetitions where available).}
  \label{fig:result03}
\end{figure}

All results are reported on the held-out Class~B exam (Sec.~\ref{sec:experiments}). We use the exam-level metrics defined in Sec.~\ref{sec:eval_metrics} (MAD, $\sigma_{\lvert\Delta\rvert}$, Bias) and, where relevant, the manual review trigger rate based on grader disagreement. Unless noted otherwise, each backend was evaluated with the full pipeline configuration and repeated $R=3$ times to estimate experimental variability; GPT-5.2-pro is reported from a single run due to cost. OpenAI models were accessed via the official OpenAI API~\cite{openai_api_ref}, while other models were accessed via OpenRouter~\cite{openrouter_api_ref}.
\paragraph{Model viability screening.}
Figure~\ref{fig:result01} compares backends under the full pipeline. The strongest backends (GPT-5.2, GPT-5.2-pro, and Gemini-3 Pro) achieve single-digit MAD with low bias, indicating close agreement with the lecturer’s exam grades. In contrast, GPT-4o and Mistral~3 exhibit substantially larger deviations; Mistral~3 also shows a pronounced positive bias, consistent with systematic over-grading on this exam.
\paragraph{Ablation on pipeline guidance.}
To isolate the effect of prompt engineering and reference conditioning, we evaluate two strong backends under a trivial prompting regime (Fig.~\ref{fig:result02}). In this setting, the prompts only enforce the Markdown output format, but omit grading rules, omit the reference-solution summary, and disable supervisor aggregation. Because aggregation is disabled, we treat the three per-grader outputs as independent grading attempts and report their mean. For both GPT-5.2 and Gemini-3 Pro, trivial prompting increases MAD and introduces a strong positive bias, confirming that structured prompting and reference grounding are necessary to obtain reliable grading.
\paragraph{Estimated manual review workload.}
Figure~\ref{fig:result03} reports the fraction of submissions that would require manual consolidation as a function of the disagreement threshold $D_{\max}$, computed at the exam level from the ensemble grader outputs. At strict thresholds (e.g., $D_{\max}=20$--30 points), weaker backends yield substantially higher review rates, reflecting less stable grading. For larger thresholds the trigger rate drops for all models, indicating that only a small subset of submissions exhibit severe grader disagreement. This analysis complements accuracy metrics by quantifying the expected human effort required to safely deploy the system.

\paragraph{Ablation on reference conditioning.}
Table~\ref{tab:ablation} ablates reference usage for the two best backends (GPT-5.2, Gemini-3 Pro) under three regimes: \emph{Full pipeline} (reference extracted to text and injected into prompts), \emph{No reference}, and \emph{Image reference} (reference image only, no text extraction). Each cell reports run1/run2/run3 and mean$\pm$std; best results per model and metric are boldfaced. We report MAD, $\sigma_{\lvert\Delta\rvert}$, Bias, and the exam-level manual review trigger rate $TR(D_{\max}=40)$ from per-grader scores. For GPT-5.2, removing the reference increases MAD and introduces a strong positive bias; image-only reference partly recovers MAD but remains biased, highlighting the role of text extraction for calibrated scoring. For Gemini-3 Pro, both ablations primarily increase positive bias with limited gains in MAD/$\sigma_{\lvert\Delta\rvert}$, consistent with pipeline development being carried out mainly on GPT-5 models, yielding a larger benefit for GPT-5.2 than for Gemini.

\begin{table*}[t]
\centering
\small
\setlength{\tabcolsep}{4pt}
\renewcommand{\arraystretch}{1.15}
\begin{tabularx}{\textwidth}{ll*{4}{>{\centering\arraybackslash}X}}
\toprule
\textbf{Model} & \textbf{Regime} & \textbf{MAD} & $\boldsymbol{\sigma_{\lvert\Delta\rvert}}$ & \textbf{Bias} & $\mathbf{TR}(D_{\max}=40)$ [\%] \\
\midrule
\multirow{3}{*}{GPT-5.2} & Full pipeline & \makecell[c]{\bfseries 7.7/8.4/7.4\\(7.8$\pm$0.4)} & \makecell[c]{\bfseries 4.2/6.5/6.9\\(5.9$\pm$1.2)} & \makecell[c]{\bfseries 0.6/-0.4/0.3\\(0.2$\pm$0.4)} & \makecell[c]{\bfseries 16.7/16.7/16.7\%\\(16.7$\pm$0.0)\%} \\
 & No reference & \makecell[c]{9.8/11.2/9.5\\(10.2$\pm$0.8)} & \makecell[c]{8.4/6.4/6.0\\(6.9$\pm$1.1)} & \makecell[c]{6.5/6.4/6.1\\(6.4$\pm$0.2)} & \makecell[c]{22.2/16.7/16.7\%\\(18.5$\pm$2.6)\%} \\
 & Image reference & \makecell[c]{7.7/8.2/8.2\\(8.1$\pm$0.2)} & \makecell[c]{5.9/4.9/7.1\\(6.0$\pm$0.9)} & \makecell[c]{3.9/3.4/4.8\\(4.0$\pm$0.5)} & \makecell[c]{\bfseries 16.7/16.7/16.7\%\\(16.7$\pm$0.0)\%} \\
\midrule
\multirow{3}{*}{Gemini-3 Pro} & Full pipeline & \makecell[c]{\bfseries 7.1/9.8/6.7\\(7.9$\pm$1.4)} & \makecell[c]{5.0/15.1/5.7\\(8.6$\pm$4.6)} & \makecell[c]{\bfseries 3.0/-2.7/0.7\\(0.3$\pm$2.3)} & \makecell[c]{\bfseries 16.7/16.7/16.7\%\\(16.7$\pm$0.0)\%} \\
 & No reference & \makecell[c]{7.4/8.9/9.6\\(8.6$\pm$0.9)} & \makecell[c]{6.8/8.2/8.6\\(7.9$\pm$0.8)} & \makecell[c]{6.6/8.4/9.0\\(8.0$\pm$1.0)} & \makecell[c]{\bfseries 16.7/16.7/16.7\%\\(16.7$\pm$0.0)\%} \\
 & Image reference & \makecell[c]{7.9/8.5/8.4\\(8.3$\pm$0.3)} & \makecell[c]{\bfseries 6.9/6.2/7.4\\(6.9$\pm$0.5)} & \makecell[c]{7.1/7.8/7.7\\(7.5$\pm$0.3)} & \makecell[c]{\bfseries 16.7/16.7/16.7\%\\(16.7$\pm$0.0)\%} \\
\bottomrule
\end{tabularx}
\caption{Ablation study on the two best-performing backends under three reference regimes: \emph{Full pipeline} (reference extracted into text and injected into prompts), \emph{No reference}, and \emph{Image reference} (reference image only, no text extraction). Each cell reports run1/run2/run3 and mean$\pm$std across runs. Metrics are exam-level MAD, $\sigma_{\lvert\Delta\rvert}$, Bias, and the manual review trigger rate $TR(D_{\max}=40)$ computed from per-grader exam scores. Best results per model and metric are highlighted in bold (lowest mean; for Bias, smallest absolute mean).}

\label{tab:ablation}
\end{table*}

\paragraph{Student feedback (Class A).}
We collected preliminary student feedback in Class~A after real use of the system on 8 weekly AI-graded quizzes. A total of 14 students completed an anonymous questionnaire; at the time of the survey, students had already received detailed PDF feedback and had access to a complaint process. Attitudes toward AI-first grading were mostly positive: 64\% reported a positive stance or preference, 21\% negative, and 14\% indifferent. When asked whether they benefit from the system, 71\% answered yes (29\% no). Reported benefits (multiple-choice) were primarily detailed explanations (41\%), perceived fairness/impartiality (34\%), and fast turnaround (21\%). The main concerns were missed answers (35\%), more mistakes than professors (26\%), and changed exam difficulty (22\% less demanding, 17\% more demanding). Overall, 43\% judged that advantages outweigh disadvantages, 29\% the opposite, and 21\% reported no difference.

\section{Discussion}
To the best of our knowledge, the literature does not yet describe a comparable end-to-end framework that grades \emph{scanned, multi-page, handwritten STEM exams with diagrams} using multimodal LLMs while producing deterministically parseable outputs with explicit guardrails, aggregation, and auditable reports. Our results show that, with such workflow design, modern multimodal backends can grade short engineering quizzes with agreement close enough to enable practical use with limited manual intervention. In informal discussions with instructors, we repeatedly encountered skepticism that this would be achievable for unconstrained handwriting and sketches at the level of accuracy reported here; these experiments provide evidence that the capability is now real when the system is engineered around model failure modes rather than idealized prompts.
A key contextual point is timing: in our experience, this type of end-to-end approach only became practically viable with the late-2025 generation of multimodal \emph{reasoning}-capable models released by major providers (e.g., GPT-5/GPT-5.2 and Gemini~3) \cite{openai_gpt5,openai_gpt52,google_gemini3}. Earlier backends in our screening exhibit substantially larger deviations and stronger bias (Fig.~\ref{fig:result01}), reinforcing the need for both capable models and systems-level safeguards.
The presented pipeline was built first for real instructional use, not as a benchmark-optimized research prototype. Accordingly, our evaluation is preliminary: results are reported on one held-out quiz with a single human grader as reference, and the underlying exam data cannot be released in its raw form due to privacy constraints. We therefore plan to expand validation to a larger and more diverse collection and, following privacy review and institutional approval, release the code, prompts, and an accompanying dataset suitable for standardized evaluation. The absence of widely usable end-to-end datasets for authentic handwritten exam grading remains a practical barrier for the field; we view this work as an initial step toward making such evaluation feasible and repeatable.

\section{Conclusion}
We presented an end-to-end workflow for grading scanned handwritten engineering exams with multimodal LLMs. The core contribution is a robust system design that couples prompt pairs and rigid templates with deterministic validation, ensemble grading, and supervisor aggregation to turn probabilistic model behavior into auditable grading artifacts. On a held-out real course quiz, state-of-the-art multimodal backends achieve close agreement with lecturer grades and manageable estimated manual-review rates, indicating that deployment is plausible for short formative assessments. We release this as preliminary evidence that automated grading of realistic handwritten STEM work is now achievable under careful workflow constraints, and we plan broader evaluation and open-sourcing (with an accompanying dataset) after further validation and privacy review.

\section*{Ethical considerations}

The ethical consideration of this research research aims to protect both study participants and future users of the proposed technology. Key ethical concerns include the handling of sensitive data (e.g., grades), the risk that participation in the experiment could affect student performance, and the power imbalance between students and instructors.

Our approach is guided by the three principles of the Belmont Report~\cite{belmont}:
\begin{enumerate}
    \item \textbf{Respect for persons}: safeguarding autonomy through dignity, agency, and informed consent;
    \item \textbf{Beneficence}: maximizing potential benefits while minimizing risks and harms; and
    \item \textbf{Justice}: ensuring equitable treatment and a fair distribution of benefits and burdens.
\end{enumerate}

In this context, a central consideration is whether the anticipated benefits justify the burdens placed on the affected population. Potential benefits to students include (i) more objective grading and (ii) higher-quality feedback from their instructors. Individualized feedback is especially valuable because it is often infeasible for instructors to provide at scale. The resulting feedback may also support improvements to course design and teaching practices. Importantly, the automated grading system did not affect students’ official course or exam grades; all assessments were graded manually as they would have been without the study.

A full-scale deployment would require formal ethical review, including evaluation of the experimental design and data-protection measures. Nonetheless, we conclude that the anticipated benefits to students outweigh the associated burdens.

\section*{Disclosure of AI use}
The whole orchestration pipeline and the experimental code (approximately 13.000 lines of python code) has been written with the help of GPT5-codex tool and GPT5-pro models by OpenAI. Search for related work was done using OpenAI's AI agent (DeepResearch), and manuscript text was written by dictating the contents to the GPT5-pro. The manuscript has been throroughly checked by the authors and revised where necessary.

\section*{Supplementary material}
\label{sec:supplementary}
We provide the scanned reference solution of the "Class B" exam, and a corresponding mock solution, which was actually graded by the pipeline using GPT-5.2 with thinking set to "high". It can be accessed via the following link: \url{https://lmi.fe.uni-lj.si/en/janez-pers-2/supplementary-material/} 

\section*{Acknowledgements}
We acknowledge the support of the EC/EuroHPC JU and the Slovenian Ministry of HESI via the project SLAIF (grant number 101254461).

{
    \small
    \bibliographystyle{ieeenat_fullname}
    \bibliography{references}
}

\end{document}